\documentclass{article}

\usepackage[main, final]{neurips_2026}
\usepackage[utf8]{inputenc}
\usepackage[T1]{fontenc}

\usepackage{hyperref}
\usepackage{url}

\usepackage{amsmath}
\usepackage{amsfonts}
\usepackage{nicefrac}

\usepackage{graphicx}
\usepackage{booktabs}
\usepackage{adjustbox}
\usepackage{multirow}
\usepackage{makecell}
\usepackage{wrapfig}
\usepackage{enumitem}
\usepackage{float}
\usepackage{comment}
\usepackage[normalem]{ulem}
\usepackage{caption}
\usepackage{subcaption}

\usepackage{xcolor}
\usepackage[table]{xcolor}

\usepackage{microtype}

\usepackage[accsupp]{axessibility}  

\usepackage{hyperref}

\usepackage{orcidlink}

\title{UTOPI: Efficient Egocentric Long-Video Understanding in AR via User-Guided Token Pre-Compression}

\author{%
  Ziqi Wang$^{1}$ \quad Su Chen$^{1}$ \quad Qiance Tang$^{1}$ \quad Jieyu Lin$^{2}$
  \tabularnewline
  \bfseries Ziyun Li$^{2}$ \quad Barbara De Salvo$^{2}$ \quad Sai Qian Zhang$^{1}$
  \tabularnewline[4pt]
  \textnormal{$^{1}$New York University \quad $^{2}$Meta Reality Labs}
  \tabularnewline
  \textnormal{\small\texttt{\textbraceleft zw4763, sc11057, qt2094, sai.zhang\textbraceright @nyu.edu}} \tabularnewline
  \textnormal{\small\texttt{\textbraceleft jyericlin, liziyun, barbarads\textbraceright @meta.com}}
}

\begin{document}

\maketitle


\begin{abstract}
As long-video understanding gains increasing attention, augmented reality (AR) devices are emerging as a compelling deployment platform. However, egocentric videos exhibit significant temporal redundancy, imposing substantial memory and computational burdens on resource-constrained AR hardware. To address this challenge, we introduce \textit{UTOPI}, a plug-and-play pre-compression module designed specifically for egocentric long videos on AR platforms. First, we leverage user motion signals to estimate inter-frame overlap and propose a viewpoint-shift-invariant pruning algorithm. Second, we exploit eye-tracking sensors to identify user fixations, enabling the separation of attended foreground content from background and its preferential retention via a fine-tuned neural network–based adjustment mechanism. Across multiple benchmarks and model backbones, \textit{UTOPI} achieves up to $95\%$ compression while preserving performance. Code and project page are available at \url{https://sai-lab-nyu.github.io/utopi/}.

\end{abstract}

\section{Introduction}
\label{sec:intro}

Augmented Reality (AR) is emerging not only as a new user interface but also as a powerful machine learning (ML) platform that integrates sensing, perception, memory, and speech. By tightly coupling the digital and physical worlds, AR enables real-time context extraction and decision making across domains such as healthcare~\cite{gerup2020augmented, chengoden2023metaverse, makinen2022user}, education~\cite{westin2022inclusive, al2023analyzing, takrouri2022ar}, and industry~\cite{chidsin2021ar}. Beyond immersive interaction, AR devices continuously generate rich multimodal data. With cameras, eye tracking, and motion sensors, they capture egocentric video together with pose and behavioral signals, creating new opportunities for long-context ML models to reason over user attention and environmental context.

As shown in Figure~\ref{fig:motivation-examples} (a), consider an AR head-mounted display (HMD) used as a memory assistant in daily activities. The device continuously captures and encodes multimodal inputs with ML models, and a brief user glance at a drawer must be reliably bound to the corresponding visual features. Later, answering a query such as “How many plates are in my drawer?” requires an episodic memory module that can index temporal embeddings, retrieve the correct event, and interface with a downstream processing engine (e.g., a large model). This capability of~\textit{long video understanding} (LVU) transforms AR from a passive sensing platform into an intelligent personal assistant with long-context reasoning, marking a critical step toward next-generation AR applications. In education, students can revisit past demonstrations to reinforce learning; in industry, recorded tasks can be replayed to identify errors; and in daily life, AR can serve as a memory aid for locating items such as keys. 

However, LVU places severe demands on resource-limited AR hardware. For example, encoding a 100-second high-definition video sampled at 1 FPS with a typical vision--language model (VLM) can require tens of gigabytes of memory, far exceeding the 8--16 GB capacity of current wearable devices~\cite{meta_quest3_specs_2026, apple_visionpro_specs_2026}. Reducing the sampling rate lowers memory usage but often degrades representation quality and downstream accuracy. Moreover, continuously running large encoder backbones for online encoding, retrieval, and reasoning is computationally prohibitive on AR devices. These constraints motivate new algorithms for efficient LVU in AR environments.


To address this challenge, we propose a \textit{perception-guided approach} for efficient LVU in AR, inspired by human attention. Human vision is nonuniform, with highest fidelity near the focus of attention and lower sensitivity in the periphery~\cite{geisler1998real,reynolds2009normalization,awh2001overlapping}. In egocentric AR video, visual content is often tied to the user’s task, and prior studies show that user intent is reflected in the questions they ask and the answers they seek~\cite{yarbus1967eyemovements,borji2014defendingyarbus,hajiabolhassani2014inverseyarbus}. Thus, unattended regions often contain spatial redundancy and contribute less to downstream reasoning. AR streams also exhibit temporal redundancy, since smooth head and body motion creates substantial overlap across consecutive frames (Figure~\ref{fig:motivation-examples}(a)). Our key idea is to exploit these spatial and temporal regularities by allocating computation and memory to salient regions and events while preserving information needed for LVU.

Building on this idea, we introduce \textit{UTOPI}, a plug-and-play  efficient egocentric Long-Video \textbf{U}nderstanding in AR via user-guided \textbf{TO}ken \textbf{P}re-compression module. UTOPI reduces spatial redundancy by using gaze-informed attention and a lightweight saliency predictor to retain important tokens in attended regions. To reduce temporal redundancy, it preserves full-fidelity representations of overlapping regions at their first appearance, estimated from AR motion signals, and replaces redundant tokens in later frames with compressed surrogate tokens. We also introduce \textit{EgoEye}, an AR-oriented LVU benchmark that reflects the \textbf{human-centric} and \textbf{attention-guided} nature of AR usage by aligning questions with users’ gaze patterns. Our contributions are summarized as follows:

\begin{itemize}

\item We introduce UTOPI, a plug-and-play user behavior–guided token pre-compression module that uses AR-native signals to compress visual tokens before the query stage, jointly reducing spatial and temporal redundancy to lower computation and memory costs.

\item By integrating UTOPI with multiple VLMs, we enable video understanding under tight AR resource constraints. We also introduce EgoEye, an egocentric video benchmark with diverse sensor signals and both attention-driven and detail-oriented QA pairs.

\item Across multiple egocentric LVU benchmarks and VLM backbones, UTOPI achieves consistent efficiency--accuracy gains, maintaining baseline accuracy while using as little as $5\%$ of visual tokens on EgoSchema and substantially reducing on-device memory cost.

\end{itemize}

\section{Background and Related Work}
\label{sec:related_works}
\begin{figure}[!t]
    \centering
    \includegraphics[width=1\linewidth]{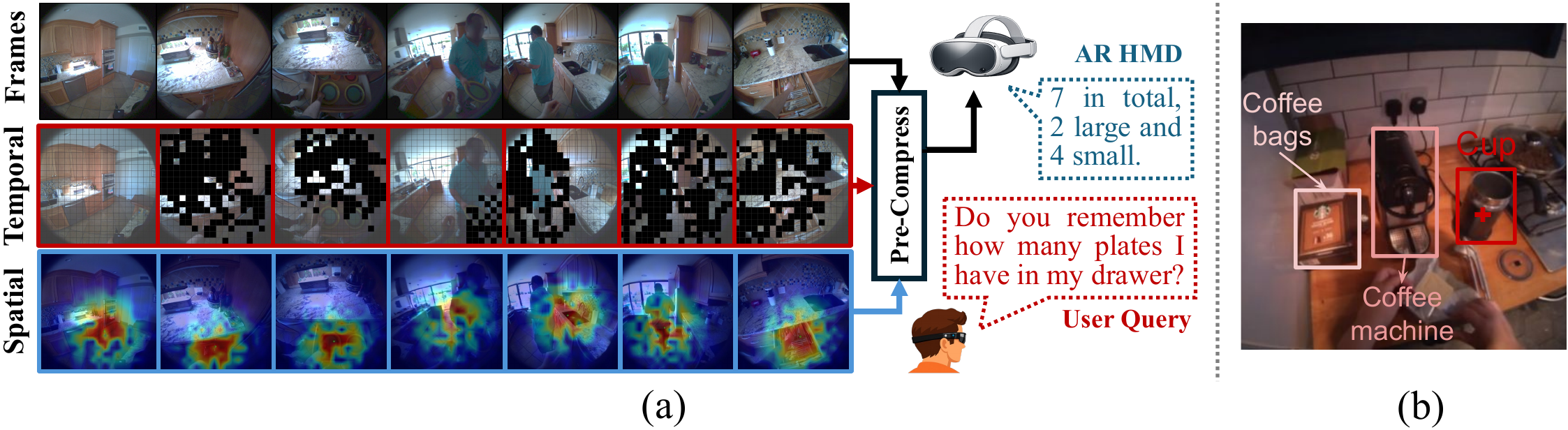}
    \vspace{-16pt}
    \caption{(a) UTOPI compresses temporal redundancy from viewpoint changes and spatial redundancy in unattended regions. (b) An example of how user attention varies with gaze.}
    \label{fig:motivation-examples}
    \vspace{-15pt}
\end{figure}

\paragraph{Efficient Long Video Understanding}
Contemporary vision--language models (VLMs), including proprietary systems~\cite{geminiteam2024gemini15unlockingmultimodal, openai2024gpt4technicalreport} and open-source models~\cite{zhang2025videollama3frontiermultimodal, deitke2024molmopixmoopenweights, bai2025qwen25vltechnicalreport}, encode images and videos into visual tokens and process them jointly with text. This unified framework supports perception, reasoning, instruction following, and dialogue, making VLMs common backbones for video understanding~\cite{maaz2023videochatgpt,zhang2023videollama,lin2023videollava,jin2023chatunivi,song2023moviechat}. Recent works have also explored gaze-guided video understanding~\cite{pham2026gazeqwen,shi2026attend}, leveraging gaze signals to improve efficiency and temporal reasoning. However, VLMs are large and memory-intensive, especially because the decoding KV cache scales with the number of visual tokens. For AR assistants, user queries often require long temporal context, further increasing memory footprint and inference latency, which makes on-device deployment difficult.

Existing efficient video methods either sample temporal frames, such as VidF4~\cite{liang2024end} and GenS~\cite{yao2025generative}, or select query-relevant spatial regions conditioned on the user question, such as DynTok~\cite{zhang2025dyntok} and Q Frame~\cite{zhang2025q}. Because these methods require the query, the video must be reprocessed for each question. In contrast, UTOPI performs query-agnostic pre-compression, allowing compressed visual tokens to be stored in the KV cache and reused across queries. Other query-free methods, such as DTD~\cite{yao2025timechat} and PruneVid~\cite{huang2025prunevid}, are not tailored to egocentric AR video and struggle with redundancy caused by continuous viewpoint changes. UTOPI instead uses user motion signals to enable viewpoint-shift-invariant token compression.

Prior work also shows that gaze is informative for modeling user behavior and intent~\cite{pani2025gazevlm,min2021integrating,lall2025eyesontarget,li2023eye}. However, these methods mainly target accuracy and may introduce costly pipelines such as optical flow~\cite{li2023eye,min2021integrating}. Other gaze-based acceleration methods~\cite{rekimoto2025gazellm,yoon2026gazevlm,chen2025eyegazecompute} often rely on simple zero-shot heuristics such as central cropping or foveated tokenization. In contrast, UTOPI uses gaze signals with a lightweight saliency gating predictor fine-tuned on EgoEye and EgoSchema, enabling more adaptive and accurate gaze-guided token compression.

\begin{wrapfigure}{r}{0.5\linewidth}
\centering
\includegraphics[width=\linewidth]{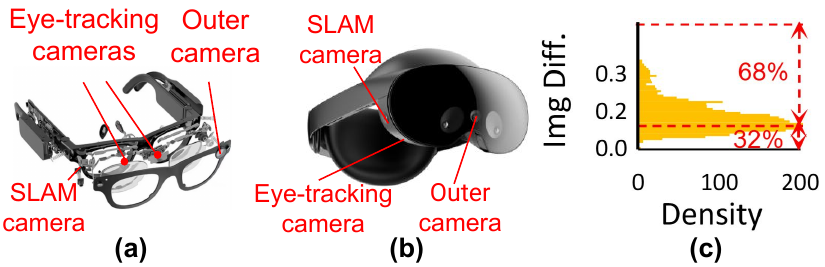}
\caption{(a) Meta Aria glasses equipped with multiple cameras. (b) Meta Quest Pro headset, which functions as both an AR and VR device. (c) Distribution of image differences.}
\label{fig:combined_figure}
\vspace{-12pt}
\end{wrapfigure}

\paragraph{Attention-Guided Question Formation in Egocentric AR}
Human perception is not uniform across the visual field but instead prioritizes attended regions. Classic studies characterize visual attention as a ``spotlight''~\cite{Eriksen1986,Posner1980}, often approximated by gaze position. Attention typically declines with increasing distance from this focus and can be modeled using functions such as a two-dimensional Gaussian~\cite{Ioannides2010}, resulting in higher fidelity near the attended location and progressively lower fidelity in the periphery~\cite{desimone1995neural,carrasco2011visual}.
This selectivity has direct implications for the types of questions an AR user is likely to ask. As illustrated in Figure~\ref{fig:motivation-examples} (b), a user wearing an AR HMD in a kitchen may fixate on a cup on the counter; objects that receive sustained attentional focus naturally become high-priority candidates for future queries, such as “Where did I leave the cup?”. In contrast, nearby objects with weaker attentional signals, shown with lighter bounding boxes in Figure~\ref{fig:motivation-examples} (b), are less likely to be referenced. From a modeling perspective, gaze thus provides a~\textbf{useful prior for identifying semantically important tokens in egocentric visual streams}. Consistent with this view, prior work in cognitive psychology and vision science shows that gaze patterns are predictive of subsequent memory formation and information-seeking behavior~\cite{yarbus2013eye,land2001ways}.

\paragraph{Human Attention in AR Environments}

To study AR users’ attention patterns over time, we use the \textit{ARIA Everyday} (AEA) dataset~\cite{lv2024aria}, collected with Meta Aria smart glasses. The dataset includes over 7.5 hours of egocentric multimodal recordings from more than 150 participants performing everyday activities, with synchronized first-person video, gaze trajectories, and head motion. As shown in Figure~\ref{fig:combined_figure}(c), we measure RGB pixel-wise differences between consecutive frames to quantify visual changes. About $32\%$ of frame pairs show less than $9\%$ change, indicating substantial temporal redundancy and strong correlation across adjacent frames. This suggests that much of the visual content in egocentric AR video remains stable over short intervals, enabling opportunities for efficient processing and computation reuse.

\paragraph{AR Hardware System}
\label{sec:bg:ArHardware}
Figure~\ref{fig:combined_figure} shows a representative AR hardware configuration. Modern headsets typically include multiple front- and side-facing cameras for capturing the surrounding scene and user state, including gaze direction. Outward-facing cameras provide high-resolution scene views, e.g., $1408 \times 1408$ on Meta quest pro~\cite{quest_pro}, while inward-facing cameras capture lower-resolution monochrome eye images. These sensors support gaze estimation~\cite{meta_quest3,hololens,quest_pro} through analytic methods such as pupil--corneal reflection or learning-based approaches~\cite{liu2025fovealnet}, providing key signals for user attention and interaction.
Tracking cameras, together with onboard IMU measurements, further enable visual--inertial odometry and full 6-DoF head-pose estimation, similar to ORB-SLAM3~\cite{campos2021orb}. IMU and depth sensors are standard in modern AR systems such as HoloLens 2~\cite{MicrosoftHoloLens2}, Apple Vision Pro~\cite{apple_vision_pro}, and Magic Leap 2~\cite{magic-leap-2}. This sensing stack provides visual input, gaze information, and pose tracking, making AR platforms suitable for intention-aware perception and interaction.

However, mainstream AR headsets typically provide only 8--16 GB of on-device memory~\cite{meta_quest3_specs_2026, apple_visionpro_specs_2026}, often shared by CPU and GPU under a unified memory architecture~\cite{meta_quest3_specs_2026, apple_visionpro_specs_2026, magic-leap-2}. Since part of this memory is reserved by the system and compute is limited by mobile-class chips, deploying LVU models on AR devices remains challenging and often leads to high multimodal response latency.

\section{Methodology}
\label{sec:methodology}
\begin{wrapfigure}{r}{0.7\linewidth}
\vspace{-12pt}
\centering
\includegraphics[width=\linewidth]{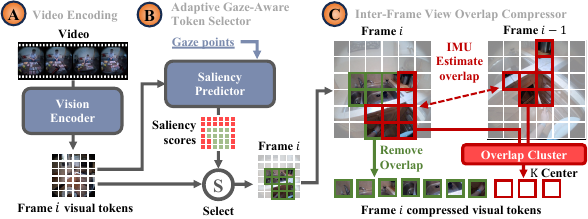}
\caption{The overall workflow of UTOPI.}
\vspace{-10pt}
\label{fig:UTOPI-AR}
\end{wrapfigure}

\subsection{UTOPI Overview}
\label{sec:UTOPI-AROverview}
VLMs process a video by first sampling frames to form a frame set $\mathbf{I}$, and then encoding each frame using a ViT-based vision encoder to produce visual tokens $\mathbf{T} = \{\mathbf{T}^i\}_{i=1}^{N}$, where
$\mathbf{T}^i = \{\mathbf{T}^i_{w,h} \mid w \in [1, W],\; h \in [1, H]\}$, $i$ denotes the frame index, $N$ is the number of sampled frames, and $(w,h)$ represents the spatial location of token $\mathbf{T}^i_{w,h}$ within frame $\mathbf{I}^i$, with an $H \times W$ token grid. Positional embeddings $P(\cdot)$ are then applied to these tokens to obtain the embedded representation $\mathbf{T}_{\mathrm{emb}} = P(\mathbf{T})$. Finally, $\mathbf{T}_{\mathrm{emb}}$ is fed together with textual tokens into the language decoder to generate the final output. To reduce computation and memory overhead on AR devices, we aim to compress the visual tokens in $\mathbf{T}_{\mathrm{emb}}$, thereby reducing the key--value (KV) cache size required for later user queries and lowering the associated decoding cost.


The overall UTOPI pipeline, shown in Figure~\ref{fig:UTOPI-AR}, consists of three stages: \textit{Video Encoding}, \textit{Adaptive Gaze-Aware Token Selector (AGTS)} and \textit{Inter-Frame View Overlap Compressor (IVOC)}. After obtaining visual tokens $\mathbf{T}$ for all frames from the video encoding stage, we first apply the AGTS to each frame $\mathbf{I}^i$. The AGTS leverages real-time gaze information together with a learnable~\textit{Saliency Predictor} to estimate token-level saliency and select foreground tokens related to the user’s attention, while discarding low-saliency background tokens. IVOC is then applied to these selected foreground tokens to further reduce temporal redundancy by using IMU-based pose estimation to identify overlapping regions between consecutive frames. Redundant overlap tokens are removed and replaced by a small set of clustered surrogate tokens, producing the final compressed visual tokens for downstream reasoning. We next describe the implementation details.

\subsection{Adaptive Gaze-Aware Token Selector (AGTS)}
\label{sec:AGTS}
According to Section~\ref{sec:related_works}, gaze location provides a strong prior for identifying semantically important tokens in egocentric visual systems. Modern AR devices are equipped with eye-tracking modules that record the 2D gaze coordinates $(g_x^i, g_y^i)$ for each frame $\mathbf{I}^i$. However, simple gaze-based subsampling, such as cropping around the gaze location, is often insufficient to preserve strong LVU accuracy. To better exploit gaze information, we use it as a prior cue for token importance and employ a neural network to refine the importance scores for improved performance.
Specifically, we apply a 2D Gaussian kernel centered at $(g_x^i, g_y^i)$ on frame $i$:
\begin{equation}
\small
G^i(w,h)=\exp\!\left(-\frac{(w-g^i_x)^2+(h-g^i_y)^2}{2\sigma^2}\right)
\end{equation}
where $\sigma$ is a predefined standard deviation. Given this prior, we further introduce a saliency predictor, implemented as a lightweight MLP, which takes each visual token as input and produces an importance score $\alpha^i_{w,h}=\operatorname{MLP}(\mathbf{T}^i_{w,h})$.
The final foreground tokens are then selected as:
\begin{equation}
\small
\textbf{T}_{F}^i=
\{\mathbf{T}^i_{w,h}\mid \alpha^i_{w,h}\cdot G^i(w,h)>\rho\}
\label{eq:gaze_foreground_selection}
\end{equation}
Here, $\rho$ is a predefined threshold, and $w,h$ denote the spatial location of token $\mathbf{T}^i_{w,h}$. The positional indices \((i,w,h)\) of each foreground token are kept identical to the original, preserving spatiotemporal structure of foreground tokens.

To better align the user gaze distribution with the LVU task, we fine-tune the MLP using a calibration dataset constructed by randomly sampling 2k QA pairs from EgoEye while freezing the remaining components of the VLM to reduce training cost. Specifically, during the forward pass, after obtaining $\mathbf{T}_{F}^i$ via Equation~\ref{eq:gaze_foreground_selection}, we compute the final loss using next-token prediction $\mathcal{L} _{NTP}= -\log \frac{\exp(\mathbf{z}_{y})}{\sum_{v=1}^{V} \exp(\mathbf{z}_{v})}$, where $\mathbf{z}=F_{dec}(\textbf{T}_{F}^i)$. $F_{\mathrm{dec}}(\cdot)$ denotes the decoder function, $\mathbf{z} \in \mathbb{R}^V$ represents the predicted logits over the token vocabulary, where $V$ is the vocabulary size, and $\mathbf{z}_y$ is the logit corresponding to the correct answer token. Since the language decoder is differentiable, we can compute $\partial \mathcal{L}_{\mathrm{NTP}}/\partial \mathbf{T}_{F}^i$.

However, $\mathbf{T}_{F}^i$ is obtained from $\mathbf{T}_{w,h}^i$ via the hard selection defined in Equation~\ref{eq:gaze_foreground_selection} and is therefore not a differentiable function of the gate values $\boldsymbol{\alpha}^i$. As a result, $\partial \mathbf{T}_{F}^i/\partial \boldsymbol{\alpha}^i$ is undefined. We thus adopt a straight-through estimator (STE) and approximate the gradient with respect to $\alpha_{w,h}^i$ as follows:
\begin{equation}
\label{eq:ste}
\small
\frac{\partial \mathbf{T}^i_{w,h}}{\partial \alpha^i_{w,h}}\approx G^i(w,h)\sum_{d=1}^{D}\mathbf{T}^i_{w,h,d}, \forall~\mathbf{T}^i_{w,h} \in \mathbf{T}^i_F\end{equation}
where $\mathbf{T}^i_{w,h} \in \mathbb{R}^{D}$ is a $D$-dimensional token embedding vector. Since $\boldsymbol{\alpha}^i$ is produced by the saliency predictor, we can backpropagate the gradients to the saliency predictor using Equation~\ref{eq:ste}.

\begin{figure}[!t]
    \centering
    \begin{minipage}[t]{0.65\linewidth}
        \centering        \includegraphics[width=\linewidth]{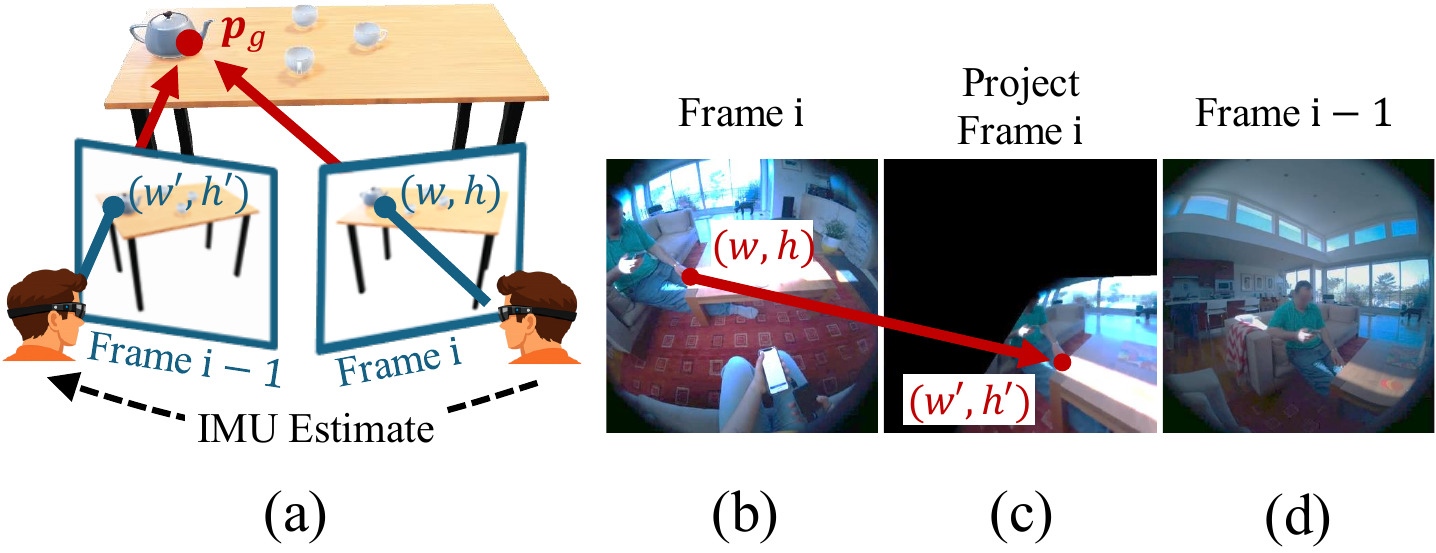}
        \vspace{-5pt}
        \captionof{figure}{IMU-based inter-frame coordinate projection.}
        \label{fig:overlap}
        \vspace{-12pt}
    \end{minipage}\hfill
    \begin{minipage}[t]{0.34\linewidth}
        \centering        \includegraphics[width=\linewidth]{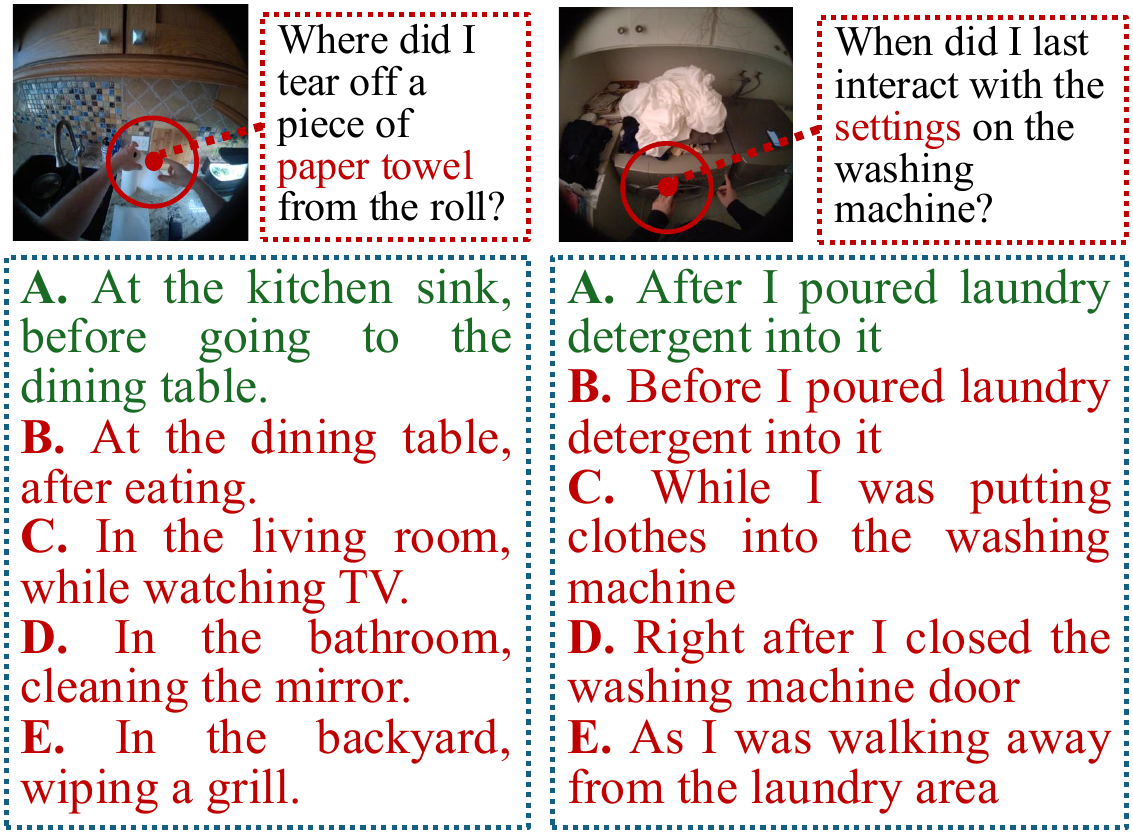}
        \vspace{-5pt}
        \captionof{figure}{EgoEye QA examples.}
        \label{fig:EEdatasample}
        \vspace{-12pt}
    \end{minipage}
   \vspace{-5pt} 
\end{figure}

\subsection{Inter-Frame View Overlap Compressor}
\label{sec:temporal_compression}
Because egocentric videos are typically captured as a continuous first-person stream, they exhibit strong scene continuity over time, as illustrated in Figure~\ref{fig:overlap}. Content observed by the user persists across frames, while the corresponding tokens often shift spatially due to changes in head pose and viewpoint. For instance, Frame $i{-}1$ and Frame $i$ in Figure~\ref{fig:overlap} illustrate two consecutive frames from the AEA dataset at times $i{-}1$ and $i$, where the man and table are visible in both frames. We observe that corresponding visual tokens generally reappear at different spatial locations rather than remaining aligned at the same position.
Although consecutive frames exhibit high temporal redundancy, directly comparing their RGB values often produces large differences. This is primarily caused by changes in the user’s head pose, which alter the camera’s orientation and position between frames. As a result, the same scene content may appear at different spatial locations in the image. Therefore, user pose information, which is naturally collected by modern AR devices, should be considered when assessing redundancy across frames.


To achieve this, we adopt a geometry-based reprojection strategy to estimate inter-frame overlaps. Specifically, for each foreground token $\mathbf{T}^i_{w,h} \in \mathbf{T}_F^i$ in the current frame $\mathbf{I}^i$ produced by the AGTS module, its camera coordinate is denoted as $(w,h)$. As shown in Figure~\ref{fig:overlap} (a), this coordinate can be mapped to a global coordinate $\mathbf{p}_g = \operatorname{proj}_{c\rightarrow g}(w,h) \in \mathbb{R}^{3}$ using the camera intrinsics.
Next, based on the camera pose change estimated from the IMU, we recover the camera pose of the previous frame $\mathbf{I}^{i-1}$ by reprojecting the global coordinate $\mathbf{p}_g$ back to that frame. This yields the corresponding camera coordinate $(w',h') = \operatorname{proj}_{g\rightarrow c}(\mathbf{p}_g)$ in $\mathbf{I}^{i-1}$. We denote the mapping of camera coordinates between two frames as $f(\cdot,\cdot)$, such that $f(w,h)=(w',h')$. If $(w',h')$ falls within the field of view of $\mathbf{I}^{i-1}$, we select the token $\mathbf{T}^{i-1}_{f(w,h)}$ in $\mathbf{I}^{i-1}$ located at this reprojected position. Specifically, we use the mapping $f(w,h)=(w',h')$ to index the corresponding token $\mathbf{T}^{i-1}_{f(w,h)}$ which represents the same visual region observed in the previous frame. Therefore, $\mathbf{T}^{i-1}_{f(w,h)}$ and $\mathbf{T}^i_{w,h}$ refer to the same physical region in the scene, observed from two consecutive frames. These corresponding regions together form the~\textbf{overlapping area} between two frames, as illustrated in Figure~\ref{fig:UTOPI-AR} (c) by the red highlighted region. The detailed mapping formulas are provided in the Appendix.
To reduce temporal redundancy, we remove redundant tokens from overlapping visual regions and represent the frames using compressed \textbf{surrogate tokens}. The process is shown in Figure~\ref{fig:UTOPI-AR} (c). First, we identify the inter-frame overlap region $\mathbf{O}^i$ between frames $\mathbf{I}^i$ and $\mathbf{I}^{i-1}$ and collect the corresponding tokens in the current frame:
$\textbf{T}_O^i=\{\mathbf{T}^i_{w,h}|(w,h) \in \textbf{O}^i,~\mathbf{T}^i_{w,h}\in\mathbf{T}_F^i\}$. Notably, \textbf{IVOC is applied only to the foreground tokens selected by AGTS}, so $\mathbf{T}_O^i \subseteq \mathbf{T}_F^i$. We then map the spatial coordinates of these tokens to the previous frame $\mathbf{I}^{i-1}$ using the mapping $f(\cdot,\cdot)$. For each token $\mathbf{T}^i_{w,h}$, we compute the cosine similarity between it and its corresponding token at the mapped location in the previous frame:
\begin{equation}
  \small
  s^i_{w,h}=\operatorname{cosim}\!\bigl(\mathbf{T}^{i-1}_{f(w,h)},\; \mathbf{T}^{i}_{w,h}\bigr),
\end{equation}
we then identify tokens within $\mathbf{O}^i$ whose representations remain sufficiently similar to those in $\mathbf{I}^{i-1}$:
\begin{equation}
  \small
  \mathbf{T}^i_S=\bigl\{\mathbf{T}^{i}_{w,h}\in \textbf{T}^i_O~|~s^i_{w,h} > \theta \bigr\},
\end{equation}
where $\theta$ is the similarity threshold. Next, we find counterparts of $\mathbf{T}^i_S$ in the previous frame $\textbf{I}^{i-1}$,
\begin{equation}
  \small
  \mathbf{D}^{i-1} = \bigl\{\mathbf{T}^{i-1}_{f(w,h)}~|~s^i_{w,h} > \theta,~(w,h)\in\mathbf{O}^i\bigr\},
\end{equation}
and cluster these tokens $\mathbf{D}^{i-1}\cup\mathbf{T}^i_S$ using k-means clustering to obtain $K$ cluster centers $\{\mathbf{T}^k_C\}_{k=1}^{K}$, as illustrated by the \textit{Overlap Cluster} module (highlighted in red) in Figure~\ref{fig:UTOPI-AR}(c), which are used as surrogate tokens to replace $\mathbf{T}^i_S$. Finally, the surrogate tokens $\{\mathbf{T}^k_C\}_{k=1}^{K}$ and the retained foreground tokens $\mathbf{T}_F^i \setminus \mathbf{T}_S^i$ are used as the compressed visual tokens for the current frame.
During subsequent VLM processing, for each surrogate token $\mathbf{T}^k_C$, we assign its temporal position to the current frame index $i$, and set its spatial position to the mean spatial coordinate of all tokens in the corresponding cluster. This coordinate is then mapped back to $\mathbf{I}^i$ via the inverse mapping $f^{-1}(\cdot,\cdot)$ and used as the final spatial location.


\subsection{EgoEye Collection}
\label{sec:egoeverything}
While multiple benchmarks have been proposed for LVU, they primarily focus on generic video-based queries and do not fully capture the \textbf{human-centric} and \textbf{attention-guided} nature of real AR usage, where questions are closely tied to user attention patterns~\cite{yarbus1967eyemovements,borji2014defendingyarbus,hajiabolhassani2014inverseyarbus}. Motivated by this gap, and building on the background presented in Section~\ref{sec:related_works}, we introduce a new benchmark, termed \textit{EgoEye}, for LVU that simulates realistic, attention-driven interactions in AR environments.

EgoEye is constructed from real-world egocentric video datasets with authentic gaze tracking, including Aria Everyday Activities (AEA)~\cite{lv2024aria} and Nymeria~\cite{ma2024nymeria}. In contrast to prior LVU benchmarks, our questions are explicitly \textbf{gaze-related}: we select question subjects by leveraging gaze traces to identify attended regions and objects, and then sample question targets with a strong bias toward items spatially close to the user’s fixation points over time. This design ensures that the queries reflect what the user actually looked at and interacted with during the experience.
EgoEye includes video clips ranging from 50 seconds to 5 minutes, with an average length of 2 minutes and 31 seconds per clip. We extract more than 2k question subjects around users' gaze locations, resulting in over 5k QA pairs. Figure~\ref{fig:EEdatasample} provides some QA examples for reference, with their question subjects being ``Paper Towel'' and ``Washing Machine Knob'', respectively. Detailed statistics and more QA examples are provided in the Appendix.

\section{Experiments}
\label{sec:experiment}
\subsection{Evaluation Setups and Implementation Details}
\label{sec:experiment:evaluation setups and implementation details}

\paragraph{\textbf{Video-QA Benchmarks}} To evaluate UTOPI, we use three Video Question–Answering (VQA) benchmarks: EgoSchema~\cite{mangalam2023egoschema}, EgoGazeVQA~\cite{peng2025eye}, and our proposed EgoEye. As UTOPI targets AR scenarios, all selected datasets are based on egocentric videos and multiple-choice questions. EgoSchema provides over 5k automatically constructed five-choice questions and offers a broad test of long-context understanding, reasoning, and retrieval. EgoGazeVQA supplies user gaze traces, with questions categorized into causal, spatial, and temporal reasoning. Both benchmarks are derived from the egocentric video dataset Ego4D~\cite{grauman2022ego4d}. Since UTOPI requires signals such as user gaze and IMU measurements, we obtain the original Ego4D metadata and align it to recover the necessary signals. 
For videos that do not have gaze signals available, we estimate the user gaze location using SUM~\cite{hosseini2025sum}, and further include an ablation study on the impact of gaze noise in Table~\ref{tab:gaze_noise}. 

Our EgoEye contains over 5k multiple-choice QA pairs (more detail in the Appendix) and includes user gaze and IMU metadata. In contrast to existing benchmarks, our questions are explicitly centered on human attention signals and realistic AR usage, producing queries that better reflect what users actually look at and care about, and enabling a more natural evaluation setting for long-context egocentric video understanding in AR environments.

\begin{wraptable}{r}{0.45\linewidth}

\vspace{-13pt}

\centering
\setlength{\tabcolsep}{3pt}
\renewcommand{\arraystretch}{0.9}

\caption{Acc. Evaluation of UTOPI.}
\label{tab:utopiar_main}

\begin{adjustbox}{width=\linewidth}

\begin{tabular}{l|c|ccc}
\toprule

\textbf{Method} &
\textbf{\makecell[c]{Token\\Budget}} &
\textbf{EE} &
\textbf{ES} &
\textbf{EGV} \\

\midrule

Qwen2.5VL-7B & 100\% & 58.4\% & 60.1\% & 47.8\% \\
Qwen2.5VL-7B & 0\%* & 36.6\% & 49.4\% & 43.4\% \\

\midrule

\multicolumn{5}{l}{\textit{Qwen2.5VL-7B}} \\

\rowcolor[HTML]{DAE8FC}
\textbf{w/ UTOPI (ours)} &
15\% &
\textbf{60.6\%} &
\textbf{64.8\%} &
\textbf{49.0\%} \\

w/ DTD~\cite{yao2025timechat} &
15\% &
57.6\% &
64.5\% &
47.9\% \\

w/ PruneVid~\cite{huang2025prunevid} &
15\% &
51.3\% &
63.0\% &
48.8\% \\

w/ Random Drop &
15\% &
56.8\% &
59.3\% &
46.1\% \\

w/ Frame Subsampling &
15\% &
48.8\% &
50.3\% &
44.6\% \\

\midrule

Qwen2.5VL-7B &
15\%$\dagger$ &
50.4\% &
53.8\% &
44.2\% \\

MindPalace~\cite{huang2025building} &
15\%$\dagger$ &
33.2\% &
45.2\% &
37.9\% \\

PLLaVA~\cite{xu2024pllava} &
15\%$\dagger$ &
34.1\% &
29.6\% &
31.5\% \\

Videollama3-7B~\cite{zhang2025videollama3frontiermultimodal} &
15\%$\dagger$ &
46.4\% &
49.7\% &
42.5\% \\

InternVL2.5-8B~\cite{chen2024expanding} &
15\%$\dagger$ &
52.4\% &
51.5\% &
43.0\% \\

\midrule

\multicolumn{5}{l}{\textit{Qwen2.5VL-7B}} \\

\rowcolor[HTML]{DAE8FC}
\textbf{w/ UTOPI (ours)} &
50\% &
\textbf{57.1\%} &
\textbf{61.9\%} &
\textbf{48.5\%} \\

w/ DTD &
50\% &
57.0\% &
60.2\% &
47.2\% \\

w/ PruneVid &
50\% &
52.4\% &
60.4\% &
47.8\% \\

w/ Random Drop &
50\% &
56.4\% &
59.6\% &
46.9\% \\

\rowcolor[HTML]{DAE8FC}
\textbf{w/ UTOPI (ours)} &
30\% &
\textbf{57.4\%} &
\textbf{62.6\%} &
\textbf{48.7\%} \\

w/ DTD &
30\% &
56.6\% &
62.1\% &
48.2\% \\

w/ PruneVid &
30\% &
53.3\% &
61.9\% &
48.1\% \\

w/ Random Drop &
30\% &
56.1\% &
60.3\% &
45.8\% \\

\bottomrule

\end{tabular}

\end{adjustbox}

\vspace{-12pt}

\end{wraptable}

\paragraph{\textbf{Implementation Details}} For video input, we set the frame resolution to $532\times532$ and uniformly sample 300 frames per video for experiments on EgoEye and EgoSchema. For EgoGazeVQA, we sample 100 frames per video, prioritizing key frames that contain gaze points. To ensure a consistent token budget across all experiments for fair comparison, we adaptively adjust hyperparameters $\theta$ and $\rho$ so that each video retains exactly the target ratio of visual tokens. 
We test the algorithms on the lightweight models including Qwen2.5VL-7B~\cite{bai2025qwen25vltechnicalreport}, Videollama3-7B~\cite{zhang2025videollama3frontiermultimodal}, PLLaVA~\cite{xu2024pllava} and InternVL2.5-8B~\cite{chen2024expanding} and use four NVIDIA A6000 (48GB) GPUs for all accuracy evaluation. Finally, we set the number of cluster centers $K$ to 2\% of the total number of tokens in the cluster, i.e., $K = 0.02 \times |\mathbf{T}^i_S|$.

To evaluate UTOPI, we compare it against other token-pruning methods for LVU tasks, including 
PruneVid~\cite{huang2025prunevid}, DTD~\cite{yao2025timechat} and MindPalace~\cite{huang2025building}. We also evaluate \textit{Random Drop} where the tokens are dropped randomly. We further evaluate \textit{Frame Subsampling}, which uniformly keeps 15\% of the sampled frames at full resolution and discards the rest. For PruneVid~\cite{huang2025prunevid}, we disable its KV-cache compression module because it requires modifications to Qwen2.5VL and some implementation details are not publicly available. For MindPalace~\cite{huang2025building}, we adopt Qwen2.5VL as the backbone LLM for generating final answers. When fine-tuning the saliency gating head, we randomly sample 2k QA pairs from EgoEye as the training set, using the remaining data for evaluation. During training, we interpret the gate values as probabilities for stochastic token dropping. We start with a low drop ratio and gradually increase it, while simultaneously annealing the temperature.

\subsection{Accuracy Performance on LVU Tasks}

In Table~\ref{tab:utopiar_main}, we evaluate several VLM baselines and combinations of Qwen2.5VL with different compression methods on three benchmarks: EgoEye (\textbf{EE}), EgoSchema (\textbf{ES}), and EgoGazeVQA (\textbf{EGV}). All methods are evaluated under identical token budgets. 
The \textit{Token Budget} in the table denotes the ratio of the number of visual tokens after compression to that before compression. A token budget marked with ``*'' indicates a value of $0\%$, meaning that only the question and answer options are provided to the model without any video input; this corresponds to a \textit{blind test}. Some token budgets are marked with ``$\dagger$'' to indicate these models do not apply token compression; instead, we downsample the video resolution to ensure that the total number of visual tokens matches ours. 

Experimental results show that under the three token budgets ($15\%, 30\%, 50\%$), UTOPI consistently outperforms all other efficiency baselines across all benchmarks. Specifically, when applied to Qwen2.5VL, UTOPI retains only $15\%$ of the visual tokens while still improving accuracy by an average of $2.7\%$ compared to using the full set of visual tokens. At the same budget, UTOPI exceeds the uncompressed model on all three benchmarks, whereas Random Drop and Frame Subsampling fall below it on all three, indicating that the gain stems from selecting informative tokens rather than from reducing the number of tokens. Compared with DTD, PruneVid, and Random Drop, UTOPI achieves $3.1\%$ higher accuracy on average. In addition, relative to naive uniform-resolution downsampling of Qwen2.5VL, UTOPI improves the average accuracy by $8.6\%$. Finally, compared with the blind test setting, UTOPI uses only $15\%$ of the visual tokens yet achieves an average accuracy gain of $15\%$, further demonstrating its effectiveness.

\subsection{UTOPI System Performance}
\label{sec:system-performance}
\paragraph{\textbf{Latency and Memory}} 
To evaluate the system performance of UTOPI, we implement the system on both a desktop GPU (RTX A6000) and an edge GPU (NVIDIA Jetson Orin NX), and report latency and peak memory usage. These metrics capture the responsiveness and resource efficiency of the system, which are critical for real-time AR workloads deployed on resource-constrained edge devices. NVIDIA Jetson Orin NX has been widely used in prior work~\cite{Pan2024CRISP, Zhang2024SubMillisecondEyeTracking, Zhang2024BOXR} to simulate AR/VR devices. This platform provides 16GB of unified memory. Since the VLMs evaluated in Table~\ref{tab:utopiar_main} cannot fully fit within the memory capacity of the edge GPU, we adopt 4-bit quantized Qwen2.5VL-7B~\cite{qwen2.5-VL} and Qwen2.5VL-3B~\cite{qwen2.5-VL} as baselines, using publicly available checkpoints from Hugging Face. We also evaluate UTOPI on SmolVLM~\cite{marafioti2025smolvlm}.



\definecolor{rowblue}{HTML}{DAE8FC}
\definecolor{hlorange}{HTML}{FAD7AC} 
\definecolor{hlred}{HTML}{F8CACA}

\newcommand{\hl}[1]{\cellcolor{hlorange}{#1}}
\newcommand{\hlr}[1]{\cellcolor{hlred}{#1}}

\begin{table}[t]
\centering
\renewcommand{\arraystretch}{1.15}

\caption{Latency and Peak Memory Evaluation. "w/ ours" indicates that UTOPI is implemented over the corresponding model.}
\label{tab:eff_mem}

\begin{adjustbox}{width=\linewidth}
\begin{tabular}{l|c|c|ccc|cc|c|c|c}
\toprule
\multicolumn{1}{c|}{\multirow{2}{*}{\textbf{Method}}} &
\multicolumn{1}{c|}{\multirow{2}{*}{\textbf{platform}}} &
\multicolumn{1}{c|}{\multirow{2}{*}{\textbf{\makecell{Token\\budget}}}} &
\multicolumn{3}{c|}{\textbf{\makecell[c]{Peak Memory Cost\\(w/o Model weights)}}} &
\multicolumn{2}{c|}{\textbf{\makecell[c]{UTOPI\\Overhead}}} &
\multicolumn{1}{c|}{\multirow{2}{*}{\textbf{\makecell[c]{TTFT\\(speed up$\times$)}}}} &
\multicolumn{1}{c|}{\multirow{2}{*}{\textbf{\makecell[c]{TPS\\(speed up$\times$)}}}} &
\multicolumn{1}{c}{\multirow{2}{*}{\textbf{EE Acc.}}} \\
\cline{4-8}
\multicolumn{1}{c|}{} &
\multicolumn{1}{c|}{} &
\multicolumn{1}{c|}{} &
\textbf{VE} & \textbf{LD} & \textbf{Overall (save\%)} &
\textbf{MFLOPs} & \textbf{Latency} &
\multicolumn{1}{c|}{} &
\multicolumn{1}{c|}{} &
\multicolumn{1}{c}{} \\
\midrule

Qwen 7B & GPU & 100\% &
6.6G & \hl{19.8G} & 19.8G &
-- & -- &
19.3s & 7.5 & 58.4\% \\

\rowcolor{rowblue}
Qwen 7B w/ ours & GPU & 15\% &
\hl{6.6G} & 4.1G & 6.6G \textbf{(67\%)} &
1636 & 673ms &
8.7s \textbf{(2.2$\times$)} & 26.6 \textbf{(3.5$\times$)} & 60.6\% \\
\hline

Qwen 7B (4bit) & Jetson Orin & 100\% &
3.2G & \hlr{7.1G} & \hlr{\textbf{OOM} 7.1G} &
-- & -- &
\hlr{\textbf{OOM} 173s} & \hlr{\textbf{OOM} 0.43} & 57.3\% \\

\rowcolor{rowblue}
Qwen 7B (4bit) w/ ours & Jetson Orin & 15\% &
\hl{3.2G} & 1.8G & 3.2G \textbf{(55\%)} &
545 & 980ms &
88s \textbf{(2.0$\times$)} & 1.3 \textbf{(3.0$\times$)} & 57.5\% \\
\hline

Qwen 3B (4bit) & Jetson Orin & 100\% &
2.9G & \hl{6.4G} & 6.4G &
-- & -- &
122s & 0.73 & 52.0\% \\

\rowcolor{rowblue}
Qwen 3B (4bit) w/ ours & Jetson Orin & 25\% &
\hl{2.9G} & 2.2G & 2.9G \textbf{(55\%)} & 545
 & 980ms &
65s \textbf{(1.9$\times$)} & 2.0 \textbf{(2.7$\times$)} & 52.1\% \\
\hline

SmolVLM 2.2B & Jetson Orin & 100\% &
2.0G & \hl{2.9G} & 2.9G &
-- & -- &
21s & 3.2 & 46.3\% \\

\rowcolor{rowblue}
SmolVLM 2.2B w/ ours & Jetson Orin & 60\% &
\hl{2.0G} & 1.8G & 2.0G \textbf{(32\%)} &
360 & 440ms &
13s \textbf{(1.6$\times$)} & 4.8 \textbf{(1.5$\times$)} & 45.9\% \\

\bottomrule
\end{tabular}
\end{adjustbox}
\end{table}

Table~\ref{tab:eff_mem} reports the \textit{Peak Memory Cost}, defined as the maximum memory usage during inference. We exclude the static memory occupied by model weights and report only the dynamic components, such as layer activations and the KV cache. \textbf{VE} denotes the peak memory cost of the Vision Encoder, while \textbf{LD} denotes that of the Language Decoder. \textbf{Overall} corresponds to the maximum of the two (highlighted in yellow). We observe that without compression, LD becomes the primary bottleneck due to the large KV cache, with memory usage on average 2.2$\times$ higher than that of the VE. With UTOPI, while maintaining comparable accuracy, the peak memory consumption is reduced by an average of 52.2\%. In particular, Qwen2.5VL-7B (4-bit) runs out of memory without UTOPI (the red entry in the corresponding row represents a reasonable estimate).


Additionally, we report latency results for UTOPI in terms of \textit{Time To First Token} (TTFT), which measures the time from input to the first generated token, and throughput, i.e., Tokens Per Second (TPS), in Table~\ref{tab:eff_mem}. Similar metrics have been used in prior work to evaluate the system performance of large models~\cite{gim2023promptcache,ye2025flashinfer,agrawal2024sarathiserve}. The results show that both response latency and decoding throughput improve substantially, achieving average gains of 1.92$\times$ and 2.67$\times$, respectively. Overall, these results demonstrate that UTOPI significantly reduces memory and latency bottlenecks while preserving accuracy, making large VLMs more practical for deployment in AR environments.

\begin{wraptable}{r}{0.32\linewidth}

\vspace{-15pt}

\centering
\setlength{\tabcolsep}{4pt}
\renewcommand{\arraystretch}{0.95}

\caption{Budget Analysis.}
\label{tab:utopiar_second}


\begin{adjustbox}{width=\linewidth}
\begin{tabular}{l|c|cc}
\toprule
\textbf{Baseline} & \textbf{\makecell[c]{Token\\ Budget}} & \textbf{EE} & \textbf{ES} \\
\midrule

PLLaVA & 100\% & \textbf{34.8\%} & 37.8\% \\
\midrule

w/ UTOPI & 50\% & 33.7\% & 39.8\% \\
w/ UTOPI & 30\% & 34.2\% & \textbf{40.1\%} \\
w/ UTOPI & 15\% & 33.5\% & \uline{39.3\%} \\
w/ UTOPI & 5\%  & 31.8\% & 32.1\% \\
\midrule

Qwen2.5VL-7B & 100\% & 58.4\% & 60.1\% \\
\midrule

w/ UTOPI & 50\% & 57.1\% & 61.9\% \\
w/ UTOPI & 30\% & 57.4\% & 62.6\% \\
w/ UTOPI & 15\% & \textbf{\uline{60.6\%}} & \textbf{64.8\%} \\
w/ UTOPI & 5\%  & 53.4\% & \uline{61.4\%} \\
w/ UTOPI & 3\%  & 50.5\% & 59.4\% \\
\bottomrule
\end{tabular}
\end{adjustbox}
\vspace{-10pt}
\end{wraptable}
\subsection{Ablation Studies}
\paragraph{Performance Under Different Token Budgets}
To evaluate the trade-off between accuracy and token budget, we apply UTOPI to PLLaVA~\cite{xu2024pllava} and Qwen2.5VL-7B and compare results under different token budgets on two benchmarks (Table~\ref{tab:utopiar_second}). A token budget of $100\%$ denotes the original model. The best accuracy is highlighted in bold, and the smallest token budget that matches or exceeds the baseline accuracy is underlined.
Results show that compressing visual tokens with UTOPI often improves accuracy on EgoSchema, likely because removing redundant tokens reduces interference. With Qwen2.5VL, our method compresses up to $95\%$ of visual tokens without accuracy loss. On PLLaVA with EgoSchema, UTOPI achieves up to $85\%$ token reduction while maintaining performance. On EgoEye, accuracy drops by less than $2\%$ at a $15\%$ token budget. Overall, UTOPI achieves a strong balance between efficiency and accuracy.

\paragraph{\textbf{Impact of AGTS and IVOC}} To validate the effectiveness of the AGTS and IVOC modules, we conduct ablation studies, as shown in Table~\ref{tab:ablation_module}. We evaluate Qwen2.5VL-7B with UTOPI under different configurations on EgoSchema. The results show that enabling either AGTS or IVOC improves accuracy to around $64\%$. When both are enabled, accuracy reaches $64.8\%$ while achieving the lowest token budget. Since AGTS and IVOC address spatial and temporal redundancies, respectively, they complement each other and jointly improve compression efficiency without harming accuracy, enabling both lower token budgets and better overall performance.

\begin{table}[t]
\centering
\scriptsize
\setlength{\tabcolsep}{3pt}
\renewcommand{\arraystretch}{1.05}

\begin{minipage}[t]{0.22\linewidth}
\centering
\caption{Evaluation of different AGTS / IVOC configurations}
\label{tab:ablation_module}

\begin{adjustbox}{max width=\linewidth}
\begin{tabular}{cc|c|c}
\toprule
\textbf{AGTS} & \textbf{IVOC} &
\textbf{\begin{tabular}[c]{@{}c@{}}Token\\Budget\end{tabular}} &
\textbf{ES} \\
\midrule
$\checkmark$ & $\checkmark$ & 15\%  & 64.8\% \\
$\checkmark$ & $\times$     & 50\%  & 64.1\% \\
$\times$     & $\checkmark$ & 30\%  & 64.4\% \\
$\times$     & $\times$     & 100\% & 60.1\% \\
\bottomrule
\end{tabular}
\end{adjustbox}

\end{minipage}
\hfill
\begin{minipage}[t]{0.52\linewidth}
\centering
\caption{Performance comparison on the Low-Motion subset “SLOW” and Motion-Intensive subset “FAST”.}
\label{tab:ablation_fast}

\begin{adjustbox}{max width=\linewidth}
\begin{tabular}{l|c|ccc}
\toprule
\multicolumn{1}{c}{\multirow{2}{*}{\textbf{Method}}} &
\multicolumn{1}{|c|}{\multirow{2}{*}{\textbf{\begin{tabular}[c]{@{}c@{}}Token\\Budget\end{tabular}}}} &
\multicolumn{3}{c}{\textbf{ES}} \\
\cline{3-5}
\multicolumn{1}{c}{} & \multicolumn{1}{|c|}{} &
\textbf{FAST} & \textbf{SLOW} & \textbf{AVG} \\
\midrule
Qwen w/ ours & 30\% & 62.0\% & 63.2\% & 62.6\% \\
Qwen w/ ours & 15\% & 63.6\% & 66.0\% & 64.8\% \\
Qwen w/ ours & 5\%  & 59.2\% & 63.6\% & 61.4\% \\
\bottomrule
\end{tabular}
\end{adjustbox}

\end{minipage}
\hfill
\begin{minipage}[t]{0.17\linewidth}
\centering
\captionsetup{justification=raggedright}
\caption{Impact of gaze noise.}
\label{tab:gaze_noise}

\begin{adjustbox}{max width=\linewidth}
\begin{tabular}{c|c|c}
\toprule
\textbf{\begin{tabular}[c]{@{}c@{}}Gaze\\Noise\end{tabular}} &
\textbf{EE} & \textbf{EGV} \\
\midrule
0 px   & 60.6\% & 49.0\% \\
75 px  & 60.3\% & 48.7\% \\
150 px & 60.4\% & 47.7\% \\
225 px & 58.9\% & 47.4\% \\
300 px & 57.7\% & 47.1\% \\
\bottomrule
\end{tabular}
\end{adjustbox}

\end{minipage}
\vspace{-12pt}
\end{table}

\paragraph{\textbf{Impact of Motion-Intensive Video}} While most videos in the test set contain egocentric footage with typical human motion, we assess IVOC’s robustness under unusual conditions by comparing Qwen2.5VL-7B with UTOPI on \textit{Motion-Intensive} and \textit{Low-Motion} videos (see Table~\ref{tab:ablation_fast}). To construct the evaluation split, we first encode each EgoSchema video with the Qwen2.5VL vision encoder and obtain a token map for every frame, denoted $\mathbf{T}^i$. We then average all token embeddings within each frame to get a single frame-level representation, $\mathbf{r}_i \;=\; \operatorname{mean}(\mathbf{T}^i)$. Next, we compute the cosine similarity between consecutive frame representations and average the similarities over the video to obtain the mean inter-frame similarity, namely
$
  \alpha \;=\; \frac{1}{N-1}\,\sum_{i=1}^{N-1}
  \operatorname{cosim}\!\bigl(\mathbf{r}_i,\mathbf{r}_{i+1}\bigr).
$
After that, we sort all EgoSchema videos by $\alpha$ and select the top $50\%$ (larger $\alpha$) as \textit{Low-Motion} videos (denoted as SLOW); the remaining $50\%$ are \textit{Motion-Intensive} videos (denoted as FAST). 

As shown in Table~\ref{tab:ablation_fast}, under the same token budget, SLOW videos generally achieve higher accuracy than FAST ones. However, the gap is small, averaging $2.6\%$, indicating that UTOPI remains robust across videos with different levels of temporal redundancy.










\paragraph{\textbf{Impact of Gaze Noise}} 
Finally, we evaluate the impact of gaze tracking noise on UTOPI. Specifically, we inject perturbations of varying magnitudes into the gaze signals in EgoEye and EgoGazeVQA, ranging from 75\,pixels to 300\,pixels. For instance, at a noise level of 75\,pixels, the perturbed gaze point is uniformly sampled within a circle centered at the ground-truth gaze point with a radius of 75\,pixels. Table~\ref{tab:gaze_noise} reports the results, where the baseline is Qwen2.5-VL-7B with a token budget of 15\%. All other settings follow Table~\ref{tab:utopiar_main}. Even under large gaze noise, the accuracy drop remains within $3\%$. These results show the robustness of our method.

Additionally, when gaze tracking is unavailable, UTOPI can still operate by estimating gaze locations with a gaze predictor, such as SUM~\cite{hosseini2025sum}, as discussed in Section~\ref{sec:experiment:evaluation setups and implementation details}. To evaluate this setting, we replace the ground-truth gaze locations in EgoEye with SUM-predicted gaze locations and repeat the accuracy evaluation. Under a $15\%$ token budget, ground-truth gaze achieves $60.6\%$ accuracy on EE, while predicted gaze achieves $59.4\%$, with a gap of less than $1.2\%$. This small difference shows that UTOPI remains effective even when direct gaze tracking is unavailable.

\section{Conclusion and Limitations}
\label{sec:conclusion}
We propose UTOPI, a plug-and-play user behavior-guided token pre-compression module in AR environments. UTOPI improves efficiency by using gaze-informed attention to discard spatially redundant visual tokens. Experiments show that UTOPI compresses up to 95\% of visual tokens on Qwen2.5VL and $85\%$ on PLLaVA while maintaining baseline accuracy. Current evaluation of UTOPI focuses on egocentric LVU benchmarks and AR simulation platforms; future work will further validate UTOPI on real-time AR hardware and more diverse deployment settings. 
\newpage

\appendix

\section{Technical appendices and supplementary material}

\subsection{EgoEye dataset}

\vspace{-10pt}
\begin{figure}[!h]
    \centering
    \includegraphics[width=\linewidth]{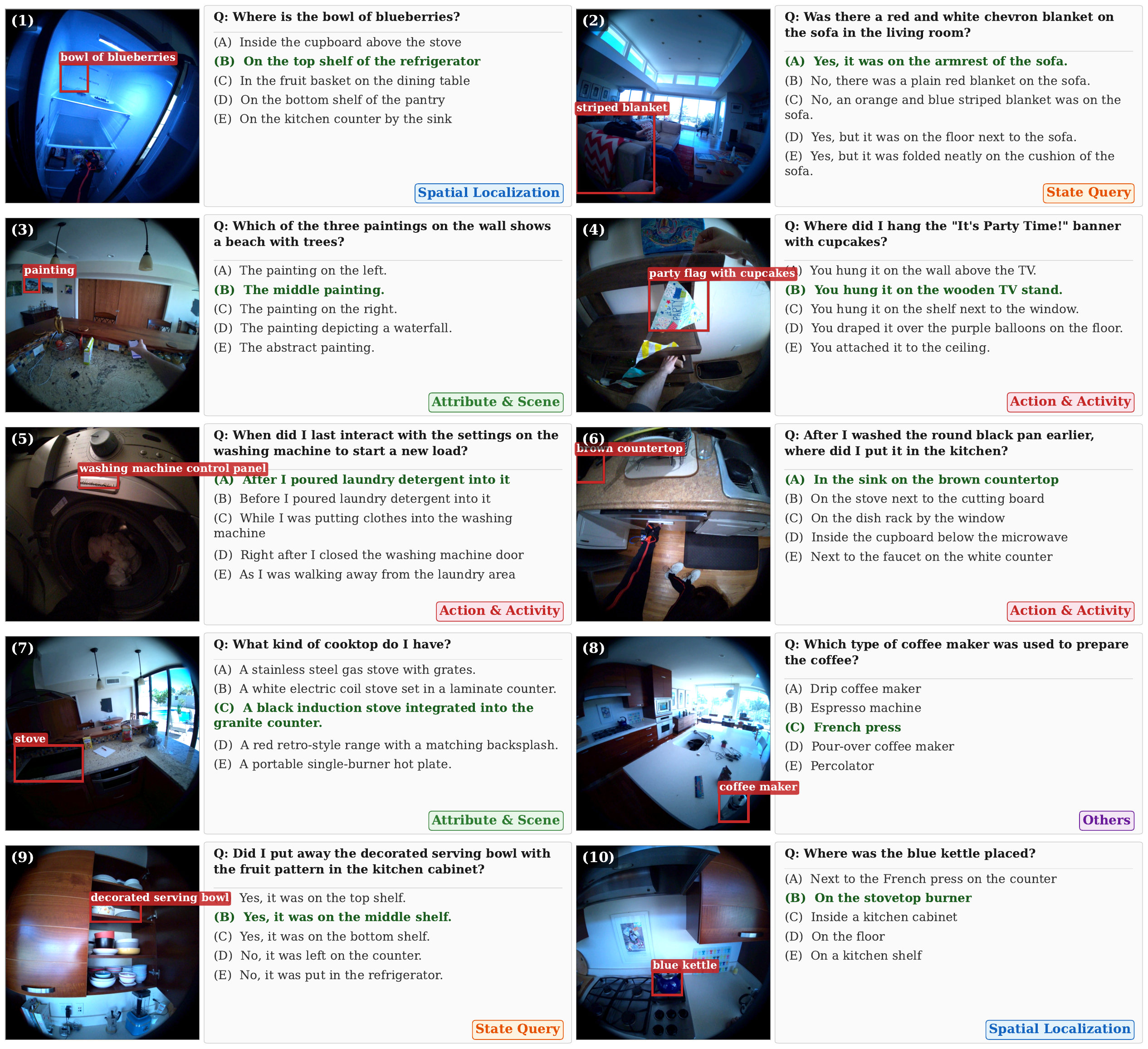}
    \caption{Additional question-option examples in EgoEye and their corresponding selected question subjects.}
    \label{fig:vqa_examples}
\vspace{-5pt}
\end{figure}

\begin{figure}[!t]
    \centering
    \begin{minipage}[t]{0.38\linewidth}
        \centering        \includegraphics[width=\linewidth]{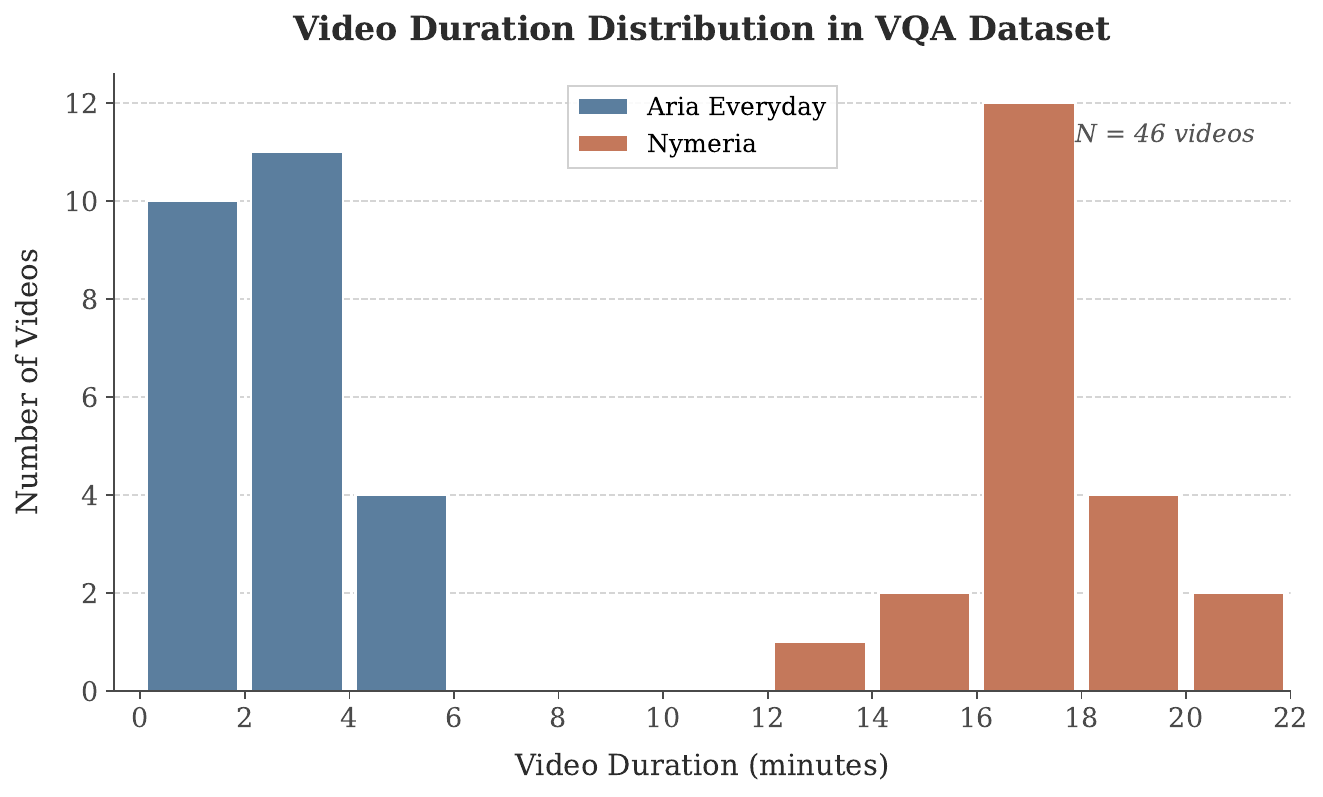}
        \vspace{-8pt}
        \captionof{figure}{Video duration distribution in EgoEye.}
        \label{fig:video_duration}
        \vspace{-12pt}
    \end{minipage}\hfill
    \begin{minipage}[t]{0.60\linewidth}
        \centering        \includegraphics[width=\linewidth]{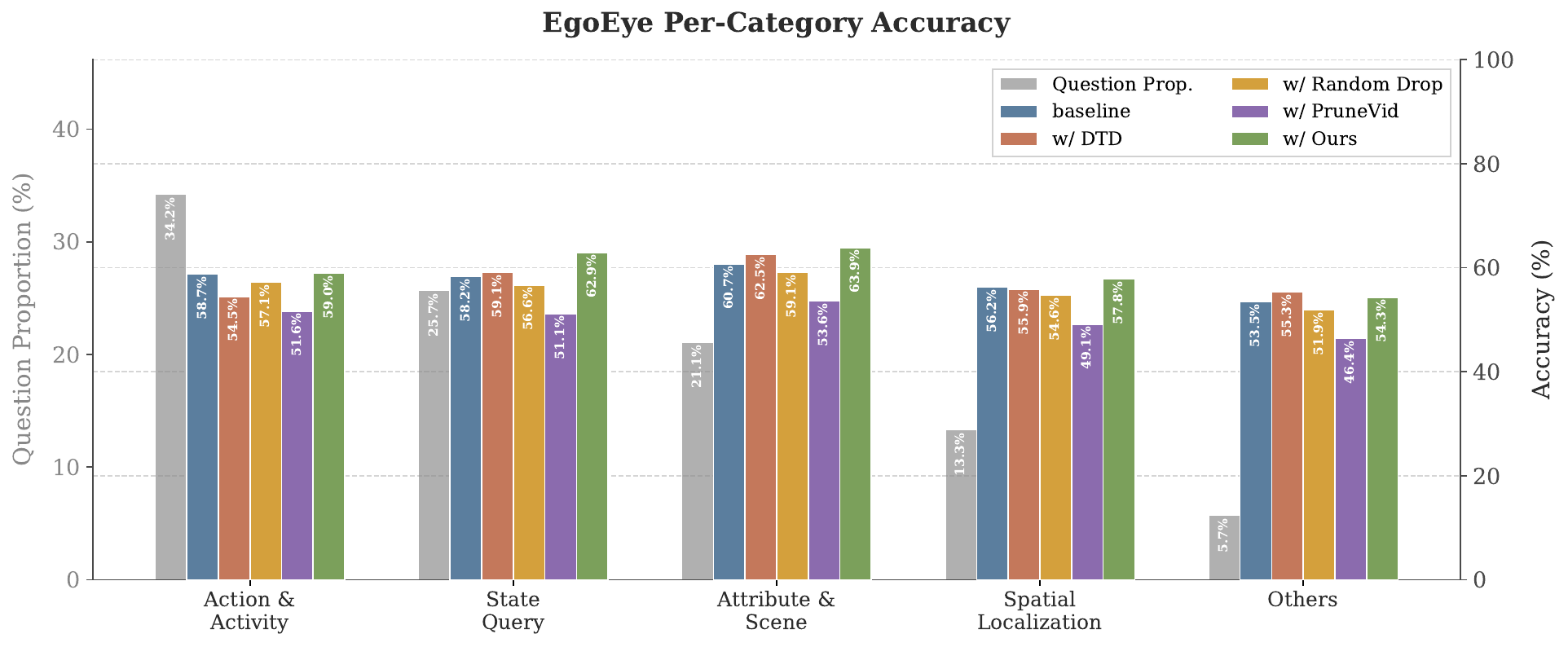}
        \vspace{-8pt}
        \captionof{figure}{EgoEye Category distribution and per-category Accuracy.}
        \label{fig:vqa_category}
        \vspace{-12pt}
    \end{minipage}
\end{figure}

The videos in EgoEye are drawn from two public datasets, Aria Everyday Activities (AEA) and Nymeria. Figure~\ref{fig:video_duration} shows the distribution of video durations in EgoEye. Based on these videos, we generated approximately 5K question-answer (QA) pairs. To construct the QA pairs, we first sample keyframes from each video and feed them, together with the user gaze coordinates at the corresponding timestamps, into a VLM. The VLM is then asked to identify an object around the gaze point as the question subject. Next, we temporally downsample the video, using a higher sampling rate for frames closer to the keyframe, and require the VLM to generate a question about the question subject, along with the corresponding options and answer. By summarizing the generated questions, we roughly group them into five high-level categories: (1) Action \& Activity, which focuses on dynamic questions such as action recognition, event ordering, and object tracking; (2) State Query, which includes yes-or-no style questions about the state of an object or scene, such as whether something is on or off, or present or absent; (3) Attribute \& Scene, which covers appearance descriptions such as color and material, as well as scene details; (4) Spatial Localization, which focuses on spatial positioning; and (5) Others, which includes remaining types of questions, such as counting. Fig.~\ref{fig:vqa_examples} provides several examples of Question-Options from each category, together with their corresponding keyframes and question subjects (highlighted by red boxes). Fig.~\ref{fig:vqa_category} shows the proportion of each category, as well as the accuracy of QWen 2.5 VL 7B with various token compression method at a 15\% token budget across different categories.

\subsection{IVOC re-projection}
\label{app:ivoc_mapping}

For a token $\mathbf{T}^i_{w,h}$ in frame $\mathbf{I}^i$, let $(u,v)$ denote the center pixel coordinate of its corresponding image patch. Here, $W_{\mathrm{img}}$ and $H_{\mathrm{img}}$ are the image width and height, while $W$ and $H$ are the width and height of the token grid. The camera intrinsic matrix is denoted by $\mathbf{K}$, where
$\mathbf{K}=\begin{bmatrix}f_x & 0 & c_x \\ 0 & f_y & c_y \\ 0 & 0 & 1\end{bmatrix}$,
with $f_x,f_y$ being the focal lengths and $(c_x,c_y)$ being the principal point. $(\mathbf{R}_i,\mathbf{t}_i)$ denotes the 6-DoF camera pose of frame $I^i$, obtained from the AR device's pose tracking system, where $\mathbf{R}_i$ and $\mathbf{t}_i$ represent the camera rotation and translation in the global coordinate system, respectively.

The corresponding global 3D coordinate is computed as
\begin{equation}
\mathbf{p}_g
=
\mathbf{R}_i
\left(
d\,\mathbf{K}^{-1}
\begin{bmatrix}
u\\
v\\
1
\end{bmatrix}
\right)
+\mathbf{t}_i.
\end{equation}

$d$ denotes the depth value at $(u,v)$. For a few videos missing IMU or depth signals, we get camera motion and depth using MegaSaM~\cite{li2025megasam} and use them as the sensor input from the AR/VR device. We then reproject $\mathbf{p}_g$ into frame $I^{i-1}$:
\begin{equation}
\begin{bmatrix}
\tilde{u}'\\
\tilde{v}'\\
\tilde{z}'
\end{bmatrix}
=
\mathbf{K}\mathbf{R}_{i-1}^{\top}(\mathbf{p}_g-\mathbf{t}_{i-1}),
\qquad
u'=\frac{\tilde{u}'}{\tilde{z}'},\quad
v'=\frac{\tilde{v}'}{\tilde{z}'}.
\end{equation}

After discretization, the mapped token index is
\begin{equation}
f(w,h)=(w',h'),
\qquad
w'=\left\lfloor u'\frac{W}{W_{\mathrm{img}}}\right\rfloor+1,\quad
h'=\left\lfloor v'\frac{H}{H_{\mathrm{img}}}\right\rfloor+1.
\end{equation}

Accordingly, the valid overlap region is
\begin{equation}
\mathbf{O}^i
=
\left\{
(w,h)\ \middle|\
\tilde{z}'>0,\ 
0\le u'<W_{\mathrm{img}},\
0\le v'<H_{\mathrm{img}}
\right\}.
\end{equation}

The inverse mapping $f^{-1}(\cdot,\cdot)$ is obtained in the same manner by swapping $i$ and $i-1$.

\subsection{Visualization}
We provide several visualization clips in Fig.~\ref{fig:vis_AGTS_IVOC} for reference. (a) shows the IVOC token drop, where the black mask indicates dropped tokens. (b) shows the AGTS saliency heat map of corresponding frame.

\subsection{Impact of Object-Gaze Distance}
\begin{figure}[t]
\centering

\begin{minipage}[t]{0.7\linewidth}
    \centering
    \includegraphics[width=\linewidth]{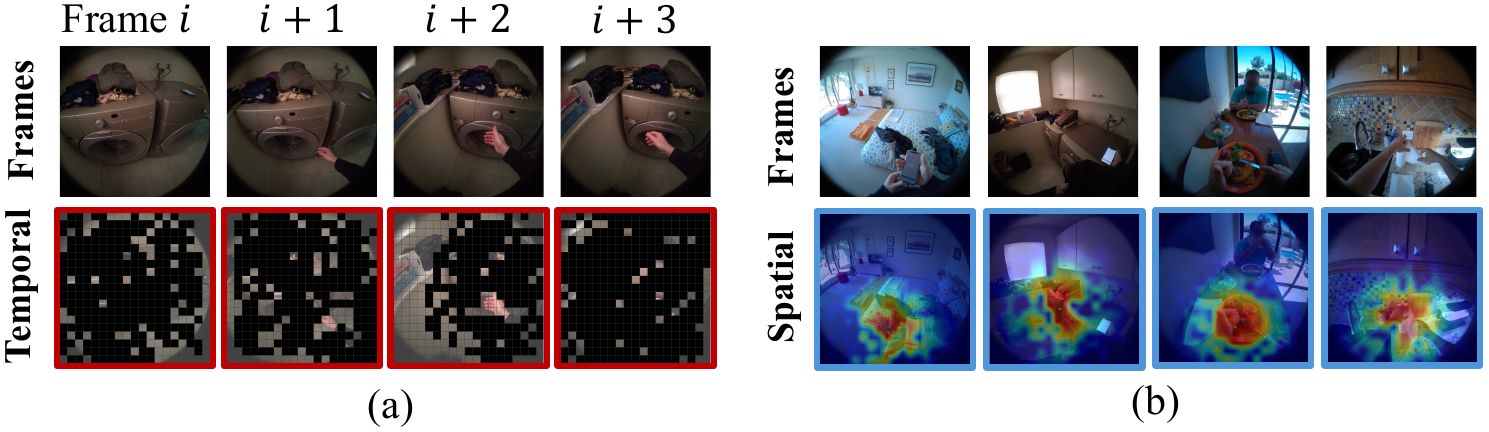}
    \vspace{-6pt}
    \captionof{figure}{Visualization of AGTS and IVOC.}
    \label{fig:vis_AGTS_IVOC}
\end{minipage}
\hspace{2pt}
\begin{minipage}[t]{0.25\linewidth}
    \centering
    \includegraphics[width=\linewidth]{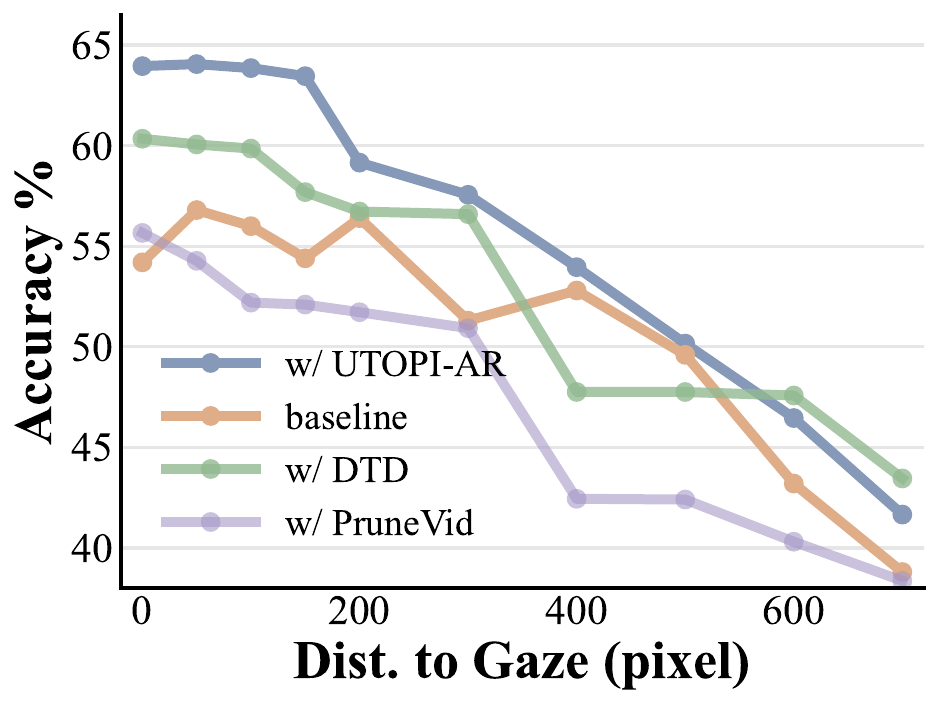}
    \vspace{-6pt}
    \captionof{figure}{Accuracy vs. question–gaze distance.}
    \label{fig:ablation_gazedist}
\end{minipage}

\vspace{-12pt}

\end{figure}

To evaluate the robustness of the AGTS, we group questions in the EgoEye benchmark by the distance between the question subject and the gaze point. We then compare the accuracy of Qwen2.5VL-7B with different compression methods across these distance groups (Figure~\ref{fig:ablation_gazedist}). Note that EgoEye frames have a resolution of $1408\times1408$. As the distance increases, all models exhibit lower accuracy, likely because objects far from the gaze point tend to appear near the frame boundary, where visual cues are less stable and questions become more challenging. When the question subject is close to the gaze point, UTOPI achieves about $5\%$ higher accuracy, demonstrating the benefit of AGTS in leveraging user attention to preserve informative regions. Even when the subject is far from the gaze point, UTOPI maintains comparable accuracy to other methods, indicating strong robustness.

\begin{figure}[h]
    \centering
    \includegraphics[width=\linewidth]{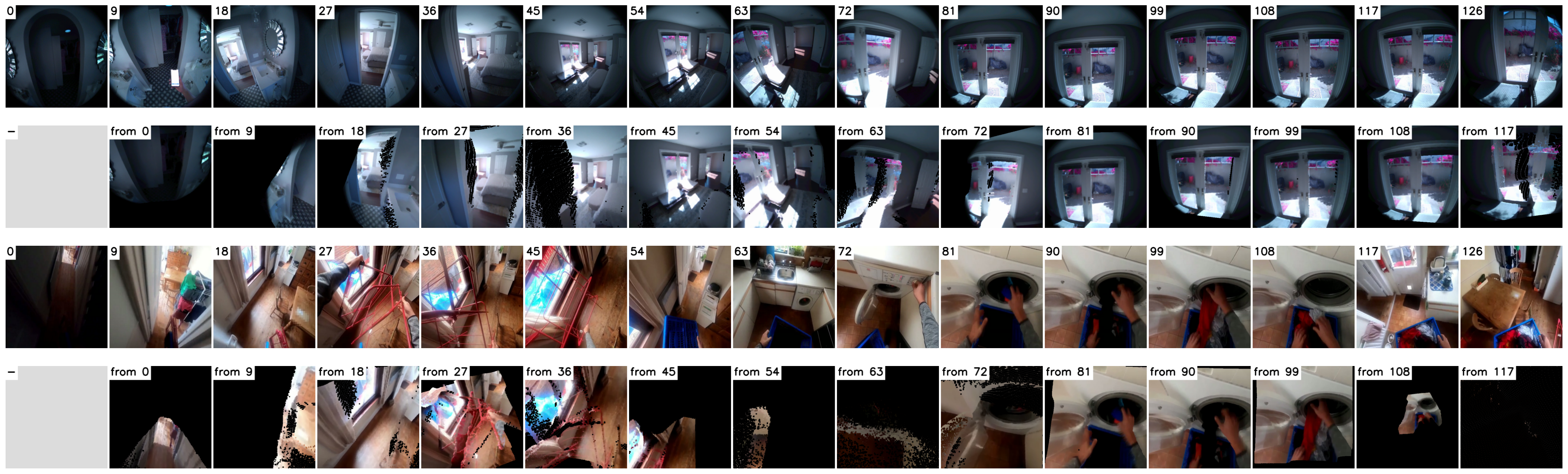}
    \caption{The visualization of re-projection. The odd rows show the original sampled frames, while the even rows show the previous sampled frames re-projected onto the current frames. To better illustrate the re-mapping quality, we present pixel-level re-projection here. As can be seen, the frames obtained by mapping from the previous frame closely match the original frames. The differences between the upper and lower frames mainly come from two factors: (1) the black regions indicate locations that are not visible in the previous frame; (2) dynamic contents, such as Frame 27 in the second video and the frame labeled “from 18”.}
    \label{fig:re-proj_example}
\end{figure}

\subsection{Broader Impacts}
\label{app:broader_impacts}
UTOPI lowers the memory and latency cost of long-video understanding on AR devices, which can support assistive uses such as memory aids, education, and error review in industry. Its intended setting, however, involves continuous egocentric recording with eye tracking. Because the retained tokens favor regions the wearer attends to, they implicitly record what the wearer looked at, and the video may also capture bystanders who have not consented to recording. Efficiency gains could further make continuous recording and analysis easier to deploy for surveillance, which UTOPI is not designed for. UTOPI performs no identity recognition and uses only 2D gaze coordinates rather than eye images. We recommend running it on device, treating compressed tokens as personal data, and giving users control over recording and retention. Finally, EgoEye is built from existing datasets and may not reflect the diversity of users, environments, and gaze behaviors in real AR use.

\subsection{Licenses of Existing Assets}
\label{app:licenses}
All datasets are used for non-commercial academic research under their respective licenses. Aria Everyday Activities~\cite{lv2024aria} is used under the Aria Everyday Activities Dataset License Agreement, and as required by that license we attribute it as ``Aria Everyday Activities Dataset, Meta Reality Labs-R''. Nymeria~\cite{ma2024nymeria} is released under the CC BY-NC 4.0 license. Ego4D~\cite{grauman2022ego4d} is used under the Ego4D License Agreement. EgoSchema~\cite{mangalam2023egoschema} is built on Ego4D videos and follows the Ego4D license terms. EgoGazeVQA~\cite{peng2025eye} is released under the CC BY 4.0 license. Model checkpoints are used under the licenses published with each release.

{
\small
\newpage
\bibliographystyle{plain}
\bibliography{ref}

}
\section*{NeurIPS Paper Checklist}

\begin{enumerate}

\item {\bf Claims}
    \item[] Question: Do the main claims made in the abstract and introduction accurately reflect the paper's contributions and scope?
    \item[] Answer: \answerYes{} 
        \item[] Justification: Section 1, 3, 5 reflect our contributions.

\item {\bf Limitations}
    \item[] Question: Does the paper discuss the limitations of the work performed by the authors?
    \item[] Answer: \answerYes{} 
    \item[] Justification: In section 5, we propose the limitations of our work.

\item {\bf Theory assumptions and proofs}
    \item[] Question: For each theoretical result, does the paper provide the full set of assumptions and a complete (and correct) proof?
    \item[] Answer: \answerNA{} 
    \item[] Justification: The paper does not present any theoretical results or formal proofs.

    \item {\bf Experimental result reproducibility}
    \item[] Question: Does the paper fully disclose all the information needed to reproduce the main experimental results of the paper to the extent that it affects the main claims and/or conclusions of the paper (regardless of whether the code and data are provided or not)?
    \item[] Answer: \answerYes{} 
    \item[] Justification: All necessary details to reproduce the main experimental results are provided in Section 4. Once the paper is accepted, we will publish an additional online report detailing all documented experimental results cited in the paper.

\item {\bf Open access to data and code}
    \item[] Question: Does the paper provide open access to the data and code, with sufficient instructions to faithfully reproduce the main experimental results, as described in supplemental material?
    \item[] Answer: \answerYes{} 
    \item[] Justification: The code, together with instructions to reproduce the main results, is released at \url{https://github.com/SAI-Lab-NYU/utopi-code}; the project page is at \url{https://sai-lab-nyu.github.io/utopi/}.

\item {\bf Experimental setting/details}
    \item[] Question: Does the paper specify all the training and test details (e.g., data splits, hyperparameters, how they were chosen, type of optimizer) necessary to understand the results?
    \item[] Answer: \answerYes{} 
    \item[] Justification: We provide the details in Section 4.

\item {\bf Experiment statistical significance}
    \item[] Question: Does the paper report error bars suitably and correctly defined or other appropriate information about the statistical significance of the experiments?
    \item[] Answer: \answerNo{} 
    \item[] Justification: We report single-run evaluation results without repeated trials or statistical significance analysis.

\item {\bf Experiments compute resources}
    \item[] Question: For each experiment, does the paper provide sufficient information on the computer resources (type of compute workers, memory, time of execution) needed to reproduce the experiments?
    \item[] Answer: \answerYes{} 
    \item[] Justification: We provide the details in Section 4.
    
\item {\bf Code of ethics}
    \item[] Question: Does the research conducted in the paper conform, in every respect, with the NeurIPS Code of Ethics \url{https://neurips.cc/public/EthicsGuidelines}?
    \item[] Answer: \answerYes{} 
    \item[] Justification: To the best of our knowledge, this work adheres to the NeurIPS Code of Ethics and complies with other relevant ethical standards.

\item {\bf Broader impacts}
    \item[] Question: Does the paper discuss both potential positive societal impacts and negative societal impacts of the work performed?
    \item[] Answer: \answerYes{} 
    \item[] Justification: We discuss positive and negative societal impacts, including wearer and bystander privacy, potential misuse, and safeguards, in Appendix~\ref{app:broader_impacts}.
    
\item {\bf Safeguards}
    \item[] Question: Does the paper describe safeguards that have been put in place for responsible release of data or models that have a high risk for misuse (e.g., pre-trained language models, image generators, or scraped datasets)?
    \item[] Answer: \answerNA{} 
    \item[] Justification: The paper does not release pretrained models or scraped data with a high risk of misuse. Deployment safeguards for systems built on UTOPI are discussed in Appendix~\ref{app:broader_impacts}.

\item {\bf Licenses for existing assets}
    \item[] Question: Are the creators or original owners of assets (e.g., code, data, models), used in the paper, properly credited and are the license and terms of use explicitly mentioned and properly respected?
    \item[] Answer: \answerYes{} 
    \item[] Justification: All datasets, models, and code used in this work are cited. The license of each dataset is listed in Appendix~\ref{app:licenses}, and model checkpoints are used under the licenses published with each release.

\item {\bf New assets}
    \item[] Question: Are new assets introduced in the paper well documented and is the documentation provided alongside the assets?
    \item[] Answer: \answerYes{} 
    \item[] Justification: This work introduces a new benchmark, EgoEye, constructed on top of existing egocentric video datasets. We provide detailed documentation of its construction process, including data generation, filtering, and evaluation protocols. We will release the dataset and accompanying documentation upon acceptance.

\item {\bf Crowdsourcing and research with human subjects}
    \item[] Question: For crowdsourcing experiments and research with human subjects, does the paper include the full text of instructions given to participants and screenshots, if applicable, as well as details about compensation (if any)? 
    \item[] Answer: \answerNA{} 
    \item[] Justification: This work does not involve crowdsourcing or human subjects research. All experiments are conducted on existing publicly available datasets, and no new data collection or annotation is performed via crowdsourcing. The manual review process in EgoEye is limited to internal quality control of generated data and does not involve external participants.

\item {\bf Institutional review board (IRB) approvals or equivalent for research with human subjects}
    \item[] Question: Does the paper describe potential risks incurred by study participants, whether such risks were disclosed to the subjects, and whether Institutional Review Board (IRB) approvals (or an equivalent approval/review based on the requirements of your country or institution) were obtained?
    \item[] Answer: \answerNA{} 
    \item[] Justification: This work does not involve any human subjects research. All experiments are conducted on existing publicly available datasets, and no new data collection or user study is performed. The manual review process is limited to quality control of generated data and does not involve interaction with human participants.

\item {\bf Declaration of LLM usage}
    \item[] Question: Does the paper describe the usage of LLMs if it is an important, original, or non-standard component of the core methods in this research? Note that if the LLM is used only for writing, editing, or formatting purposes and does \emph{not} impact the core methodology, scientific rigor, or originality of the research, declaration is not required.
    \item[] Answer: \answerNA{} 
    \item[] Justification: This work uses LLM tools solely for editing purposes, such as checking grammar or rewriting certain content and sentences within the work.

\end{enumerate}

\end{document}